\documentclass{article}

     \usepackage[preprint,nonatbib]{neurips_2019}

\usepackage[utf8]{inputenc} 
\usepackage[T1]{fontenc}    
\usepackage{hyperref}       
\usepackage{url}            
\usepackage{booktabs}       
\usepackage{amsfonts}       
\usepackage{nicefrac}       
\usepackage{microtype}      
\usepackage{graphicx}
\usepackage{amsmath}
\usepackage{bbm, dsfont}
\title{State of Compact Architecture Search For Deep Neural Networks}

\author{\hspace{-0.5in}
  Mohammad Javad Shafiee$^{1,2}$, Andrew Hryniowski$^{1,2}$, Francis Li$^2$, Zhong Qiu Lin$^{1,2}$, and~Alexander~Wong$^{1,2}$\\
  $^{1}$Waterloo Artificial Intelligence Institute, University of Waterloo, Waterloo, ON, Canada\\
  $^{2}$DarwinAI Corp., Waterloo, ON, Canada\\
}

\begin{document}

\maketitle

\begin{abstract}
The design of compact deep neural networks is a crucial task to enable widespread adoption of deep neural networks in the real-world, particularly for edge and mobile scenarios.  Due to the time-consuming and challenging nature of manually designing compact deep neural networks, there has been significant recent research interest into algorithms that automatically search for compact network architectures.  A particularly interesting class of compact architecture search algorithms are those that are guided by baseline network architectures. Such algorithms have been shown to be significantly more computationally efficient than unguided methods.  In this study, we explore the current state of compact architecture search for deep neural networks through both theoretical and empirical analysis of four different state-of-the-art compact architecture search algorithms: i) group lasso regularization, ii) variational dropout, iii) MorphNet, and iv) Generative Synthesis.  We examine these methods in detail based on a number of different factors such as efficiency, effectiveness, and scalability.  Furthermore, empirical evaluations are conducted to compare the efficacy of these compact architecture search algorithms at producing deep neural networks that have a balanced trade-off between size, speed, and accuracy across three well-known benchmark datasets.  While by no means an exhaustive exploration, we hope that this study helps provide insights into the interesting state of this relatively new area of research in terms of diversity and real, tangible gains already achieved in architecture design improvements.  Furthermore, the hope is that this study would help in pushing the conversation forward towards a deeper theoretical and empirical understanding where the research community currently stands in the landscape of compact architecture search for deep neural networks, and the practical challenges and considerations in leveraging such approaches for the purpose of aiding the design and development of deep neural networks for operational usage.

\end{abstract}
\vspace{-0.15in}
\section{Introduction}
\vspace{-0.15in}
Designing deep neural network architectures for different purposes has been an active field of research since the inception of deep neural networks. The design process has either focused on achieving better modeling accuracy or constructing compact network architectures for lower computational and architectural complexity for reduced storage and faster inference speed. Most of the well-known network architectures in literature were designed manually based on the human insights. However, automatic network architecture search and parameter tuning has recently gained more interest in the research community, due to the fact that manually designing specialized network architectures, particularly compact architectures, is very time-consuming and human insight about optimized macro- and micro-architectures is limited in terms of design granularity.

A number of algorithmic approaches have been proposed in literature for designing neural network architectures. The optimal brain damage approach proposed by Lecun {\it et al.} can be considered as the seminal work on designing compact neural network architecture, where information theory is leveraged to select non-important parameters in the model to be removed. Following that seminal work, several methods were proposed to address the storage limitations of deep neural networks.  These methods take advantage of a combination of pruning, quantization and coding techniques to address the storage requirement of a deep neural network.  Low-rank matrix factorization is another technique which was applied to approximate the filter structures and convolutional kernels in a deep neural network. Structural learning approach during the training of a model is another trick to learn the filter shape and depth during the training process. Compact architectural search was also addressed via variational Bayesian algorithms.  There have also been several studies exploring the use of evolutionary computing methods for training and generating deep neural networks~\cite{stanley2002evolving} but mostly on for reaching to better modeling accuracy.
Shafiee~{\it et~al.}~\cite{javad2016evonet} formulated the compact network architectural search problem via an evolutionary synthesis framework so-called EvoNet.

The neural architecture search (NAS) algorithms~\cite{baker2016designing} which recently gains interest from the community provides state-of-the-art modeling accuracy in pubic benchmark datasets. The search strategy in these approaches can be formulated via evolutionary algorithms,  reinforcement learning methods, or Bayesian optimization. These methods are mainly focus on finding a network architecture which provides the best modeling accuracy.

In this study, we explore the current state of compact architecture search for deep neural networks by conducting both theoretical and empirical analysis on four different state-of-the-art compact architecture search algorithms across three different datasets using accuracy, computational complexity, and architectural complexity as key performance indicators.

\vspace{-0.1in}
\section{Theoretical Analysis}
\vspace{-0.15in}
\label{sec:Method}
A key criteria when selecting the set of algorithms to study in detail for a better understanding of the current state of compact architecture search is scalability. As a result, we focus on exploring state-of-the-art compact architecture search algorithms that, guided by baseline network architectures, produce compact neural network architectures with smaller yet structured topologies that tend well to parallel computing on modern processor architectures as well as provide lower effective memory footprint during inference. Furthermore, such approaches have been demonstrated to scale well in producing compact deep neural networks that possess strong modelling accuracy for more complex problems and data.
\vspace{-0.1in}
\subsection{Group Lasso Regularization}
\vspace{-0.15in}
Group Lasso can be used to learn structured sparsity in an efficient way. For compact architecture search purposes, it was applied~\cite{wen2016learning} to regularize network structures such as filters or channels during training via the following loss function:
\vspace{-0.15in}
\begin{align}
    \label{eq:groupLasso}
    \mathcal{L}(W) = E_D(W) + \lambda_g \sum_{l=1}^L R_g(W^{(l)})
\end{align}
where $W$ represents the set of weights in the network and $E_D(\cdot)$ is the loss of the network on the data $D$. $R_g(\cdot)$ formulates the group lasso for each layer which enforces the structure sparsity during the training and $\lambda_g$ is the regularization factor.

\subsection{Variational Dropout}
\vspace{-0.15in}
The variational dropout technique, adds a Gaussian stochastic factor on each activation during the training step and the mean and standard deviation of this distribution are learned during the training. The values of these two parameters reach to zero for those weights that have no impact on the network output which promotes the sparsity. Therefore, the weights of the resulted network at the end of training are sparse. However the produced sparsity is unstructured and cannot provide any speed up inference. Here we extend upon this approach and drop those filters (i.e., a set of weights in a layer which produces one activation channel at the output) which are more sparse than a pre-defined threshold to generate the final network architecture. Setting up proper parameters and initialization of the parameters is very important to produce the best results.

\subsection{MorphNet}
\vspace{-0.15in}
MorphNet~\cite{gordon2018morphnet} performs an iterative shrink-and-expand approach to automatically design the structure of a deep neural network. It utilizes weight regularization on network activations to sparsify the network in the shrinking step, followed by a uniform multiplicative approach on all layers in the expand step.  More specifically, a penalty term is applied on the loss function during the training in the shrinking step as follows:
\vspace{-0.15in}
\begin{align}
    \mathcal{L}(W) = E_D(W) + \lambda \mathcal{F}(O_{1:M})  \;\;s.t.\;\;  &\mathcal{F}(O_{1:M}) = \sum_{L=1}^{M+1} \mathcal{F}(\text{layer  } L).
\end{align}
The $\mathcal{F}(\cdot)$ is a regularizer on neurons which can induce some of the neurons to be zeroed out and can be formulated as:
\vspace{-0.15in}
\begin{align}
    \mathcal{F}(\text{layer  } L) = C \sum_{i=0}^{I_L-1} A_{L,j} \sum_{j = 0 }^{O_L-1} B_{L,j}
\end{align}
where $A_{L,j}$ and $B_{L,j}$ are indicator functions that encode whether the input $i$ or output $j$ are alive or zeroed out. $L_1$ norm on variables of batch normalization is utilized to provide tractable learning via gradient descent. The batch-norm is applied to each layer which means that each neuron has a particular variable in the batch-norm; setting this variable to zero effectively disables the neuron. The MorphNet method needs hyper-parameter tuning; Tuning the parameters in this method plays a vital rule to produce reasonable results.
\vspace{-0.1in}
\subsection{Generative Synthesis}
\vspace{-0.15in}
The overall goal of the Generative Synthesis method~\cite{wong2018ferminets} as a compact architecture search method is to learn a generator $\mathcal{G}$ that can synthesize deep neural networks $\left\{N_s|s \in S\right\}$, given a seed set $S$, maximize a universal performance function $\mathcal{U}$ while satisfying quantitative human-specified design constraints and performance targets, as defined by an indicator function $\mathbbm{1}_r(\cdot)$.  This learning of generative machines for synthesizing deep neural networks can be formulated as the following constrained optimization problem:
\vspace{-0.05in}
\begin{equation}
\mathcal{G}  = \max_{\mathcal{G}}~\mathcal{U}\Big(\mathcal{G}(s)\Big)~~\textrm{subject~to}~~\mathbbm{1}_r\Big(\mathcal{G}(s)\Big)=1,~~\forall s \in S.
\label{optimization}
\end{equation}
\vspace{-0.1in}

Given the intractability of solving Eq. \eqref{optimization}, an approximate solution $\mathcal{G}$ to the constrained optimization problem posed is achieved by leveraging the interplay between a generator-inquisitor pair that work together in a synergistic manner to obtain not only improved insights about deep neural networks as well as learn to synthesize compact deep neural networks in a cyclical and iterative manner. A generator is first learned based on a user network design prototype, data, and human-specified design and operational requirements (size, accuracy tolerance, performance targets, hardware-level requirements such as channel multiplicity and data precision, etc.) and is used to synthesize deep neural networks.  An inquisitor probes the synthesized deep neural network and the corresponding reactionary responses are observed.  Based on these observations, the inquisitor learns at a foundational level about the architectural efficiencies of the synthesized deep neural networks, and updates the generator based on the insights it gains.  This leads to an improved generator that is then used to synthesize a new, more compact deep neural network.  The aforementioned process of synthesizing, probing, observing, and updating is repeated over cycles in an iterative fashion, resulting in a sequence of improving generators that can progressively synthesize progressively more compact deep neural networks until performance targets and operational requirements are satisfied.

\vspace{-0.1in}
\section{Empirical Analysis}
\vspace{-0.15in}
\label{sec:res}

In this study, we further examine in an empirical manner the current state of compact architecture search algorithms for producing deep neural networks. The trade-offs between size, speed, and accuracy are measured by conducting several empirical experiments using different well-known benchmark datasets.  Based on the quantitative experimental results, we study the different performance characteristics of the compact deep neural networks produced by the tested compact architecture search algorithms to gain better insights into the effectiveness as well as the search behaviours of the tested algorithms.

\vspace{-0.05in}
\subsection{Datasets \& Baseine Network Architectures}
\vspace{-0.15in}
The four state-of-the-art compact network architecture search methods studied here are evaluated using CIFAR-10, CIFAR-100~\cite{krizhevsky2009learning}, and ImageNet 64x64~\cite{deng2009imagenet}, a downsampled variant of ImageNet with over 1.28M training images across 1000 classes.

In this study, a 32 layer ResNet architecture is used as the baseline architecture for CIFAR-10 while a deeper 50 layer ResNet architecture is used for CIFAR-100 given that they are more difficult tasks.  For ImageNet 64x64, two baseline architectures were used for evaluation: i) ResNet-50, and ii) InceptionV3.

For the CIFAR-10, CIFAR-100 and ImageNet 64$\times$64 evaluations, the FLOPs target for all four of the tested methods is set to one-third of the respective baseline network architectures to quantitatively investigate the efficacy of the four tested methods at producing highly compact deep neural networks with low computational costs.

\vspace{-0.05in}
\subsection{Experimental Results \& Discussion}
\vspace{-0.15in}
The experimental results for CIFAR-10 and CIFAR-100 are shown in Table~\ref{tab:cifar10}, while the results for ImageNet 64x64 are shown in Table~\ref{tab:imagenet64}.

\textbf{CIFAR-10/CIFAR-100:}
As seen in Table~\ref{tab:cifar10}, the compact deep neural networks produced by each of the tested compact architecture search algorithms have drastically different neural network architectures with very different performance tradeoffs to meet the $\sim$48 and $\sim$74 MFLOPs ($1/3$ of the baseline) performance target for CIFAR-10 and CIFAR-100.

Conducting a close examination of the performance characteristics of the compact deep neural networks produced using the four tested compact architecture search algorithms allowed for the observation of several interesting insights:
\begin{itemize}
    \item While all produced networks have similar FLOPs as a result of the FLOPs target, several of the produced neural networks have significant more parameters than others.  For CIFAR-10, the deep neural network produced via the Group Lasso approach had noticeably more parameters than the other three methods (e.g., $\sim$33\% higher than Variational Dropout).  For CIFAR-100, the deep neural networks produced by MorphNet and Variational Dropout have more than twice the number of parameters as that produced using Group Lasso and Generative Synthesis.
    \item While all tested methods have the same FLOPs performance target ($1/3$ FLOPs of baseline), none of the methods produce networks that hit the target exactly. The deep neural network produced by Generative Synthesis had the lowest number of parameters (by around 3\% and 7\% lower than others for CIFAR-10 and CIFAR-100, respectively).
    \item Of the methods, Variational Dropout produced the neural network with the lowest modeling accuracy.  MorphNet and Generative Synthesis produced networks with better modeling accuracy than the Group Lasso and Variational Dropout methods, with Generative Synthesis producing the network with the highest accuracy amongst the tested methods for the given FLOPs target.  In fact, the neural network produced by Generative Synthesis achieved the same accraucy as the baseline architecture in the CIFAR-10 case. which illustrates that it was able to find the most balanced trade-off between size, speed, and accuracy.
\end{itemize}

\begin{table}
\setlength{\tabcolsep}{2pt}
    \caption{Results for CIFAR-10 and CIFAR-100 across four tested algorithms. Best results in \textbf{bold}.}
    \vspace{-0.1in}
    \centering
    \begin{tabular} { l | c c c | c c c }
        \toprule
        ~&\multicolumn{3}{c}{\bf CIFAR-10}& \multicolumn{3}{c}{\bf CIFAR-100} \\
        \textbf{Methods} &  \textbf{FLOPs} & \textbf{\#Parameters} & \textbf{Top-1(\%)} &  \textbf{FLOPs} & \textbf{\#Parameters} & \textbf{Top-1(\%)} \\
        \midrule

        \textbf{Baseline}
        & 144.369 & 487,754 & 91.6 & 223.946 & 769,476 & 67.8\\
                      \midrule
        \textbf{Group Lasso~\cite{wen2016learning}}

        & 46.282 & 225,564 & 86.7 & 74.852 & 268,609 & 61.6
        \\
        \textbf{MorphNet~\cite{gordon2018morphnet}}

        & 46.280 & 153,179 & 88.8 & 74.995 & 435,396 & 63.3 \\
        \textbf{Variational Dropout~\cite{molchanov2017variational}}

        & 46.046 & \textbf{151,238} & 84.9 & 74.410 &   563,197 & 57.9\\

        \textbf{Generative Synthesis~\cite{wong2018ferminets}}
& \textbf{44.658} &   152,668 & \textbf{91.6}  & \textbf{70.800} & \textbf{214,692} & \textbf{64.8}
    \end{tabular}
    \label{tab:cifar10}
\end{table}

\begin{table}
\setlength{\tabcolsep}{2pt}
    \caption{Results for ImageNet 64x64 using two baseline architectures. Best results in \textbf{bold}.}
    \vspace{-0.1in}
    \centering
    \begin{tabular} { l | c c c | c c c }
        \toprule
        \textbf{Architectures}&\multicolumn{3}{c}{\bf ResNet-50}
        &\multicolumn{3}{c}{\bf InceptionV3 }\\
        \textbf{Methods} &  \textbf{FLOPs} & \textbf{\#Parameters} &
        \textbf{Top-1(\%)} &  \textbf{FLOPs} & \textbf{\#Parameters} &
        \textbf{Top-1(\%)} \\
        \midrule
        \textbf{Baseline}
        & 2,629.21 & 25,593,400 & 60.0 & 7,515.7 & 22,895,000 & 61.8\\
        \midrule
        \textbf{Group Lasso~\cite{wen2016learning}}
        & 896.71 & 9,079,334 & 50.6 & 2506.7 & 8,160,687 & 57.6
        \\
        \textbf{MorphNet~\cite{gordon2018morphnet}}
        & 824.91 & 9,294,711 & 52.9 & 2490.6 & 9,489,201 & 58.5
        \\
        \textbf{Variational Dropout~\cite{molchanov2017variational}}
        & 873.29 & 8,850,781 & 43.3 & 2,508.0 & 8,160,687 & 53.9	
        \\
        \textbf{Generative Synthesis~\cite{wong2018ferminets}}
        & \textbf{802.14} & \textbf{3,293,420} & \textbf{57.8} &
        \textbf{2,340.3} & \textbf{5,257,960} & \textbf{62.3}
    \end{tabular}
    \label{tab:imagenet64}
\end{table}

\textbf{ImageNet 64x64:}
As seen in Table~\ref{tab:imagenet64}, similar to the \mbox{CIFAR-10/CIFAR-100} experiments, the compact deep neural networks produced by each of the tested compact architecture search algorithms produce drastically different neural network architectures with very different performance tradeoffs to meet the $\sim$876 MFLOPs and $\sim$2505 MFLOPs ($1/3$ of the baseline) performance targets for ResNet-50 and InceptionV3, respectively.

Experimental results show  several interesting insights regarding  the performance characteristics of the compact deep neural networks produced using the four tested compact architecture search algorithms:
\begin{itemize}
    \item The MorphNet approach produced the largest networks for both cases (in terms of number of parameters).   For example, MorphNet produced a network architecture that has $\sim$2.8$\times$ the number of parameters as Generative Synthesis for the ResNet-50 case.
    \item While all methods have the same FLOPs performance target ($1/3$ FLOPs of baseline), results demonstrate that, as with the previous experiment, none of the methods produce networks that hit the target exactly. However, Generative Synthesis was able to achieve lower than expected FLOPs (by $\sim$7\% lower for both cases) in this experiment as well.
    \item As with CIFAR-10 and CIFAR-100, MorphNet and Generative Synthesis produced networks with better modeling accuracy than the Group Lasso and Variational Dropout methods.  In both cases, Generative Synthesis produced neural networks that had the highest accuracy and lowest number of parameters amongst the tested methods for the given FLOPs target (e.g., $\sim$4.9\% higher accuracy than MorphNet in the ResNet-50 case) on ImageNet 64x64.  In fact, the neural network produced by Generative Synthesis outperformed the baseline architecture in terms of accuracy by 0.5\% in the InceptionV3 case. This illustrates that it was able to find the most balanced trade-off between size, speed, and accuracy.
\end{itemize}

\section{Practical Considerations}
\vspace{-0.15in}
In order to take advantage of the aforementioned compact architecture search algorithms, for the purpose of designing compact deep neural network architectures, there are a number of practical challenges unique to each method that must be considered to achieve good performance in the resulting deep neural networks.

One of the biggest challenges to effectively leveraging the Group Lasso regularization technique as a compact architecture search algorithm is to identify the optimal set of hyperparameters, with one of the most crucial hyperparameters being a proper and optimal regularization factor $\lambda_g$ during the training process. Much like Group Lasso, Variational Dropout also suffers from the \textit{curse of hyper-parameterization}. Selecting the correct regularization strength is key for allowing the model to learn. Too high a rate and a model will not learn at all, too low a rate and the model is not fully learning what parameters should be dropped.

MorphNet follows an iterative process of sparsifying a network and linearly growing the network back up. The hyper-parameters that control each step in a given iteration can differ from previous iteration. Selecting the correct rate to sparsify a network is performed via trial and error. Determining when to stop sparsifying a network during any given iteration can be unclear.

Generative Synthesis employs an iterative process to learn to generate compact deep neural networks that meet operational requirements and desired performance targets. As such, while Generative Synthesis will iterate until it hits the desired performance targets, a practical challenge with Generative Synthesis is that the number of iterations can vary depending on the complexity of the neural network architecture, the desired performance targets, as well as the complexity of the underlying data.

 \section{Conclusions}
 \vspace{-0.15in}
In this study, we explore in detail the current state of compact architecture search for deep neural networks both theoretically and empirically, as well as studying practial considerations when leverage the current state-of-the-art.  In summary, from both theoretical and empirical perspectives, the current state of compact architecture search is quite interesting and diverse, and at a stage where real significant gains can be obtained for accelerating the design of deep neural networks with compact architectures for operational usage.  In particular, it was found that purely regularization-driven compact architecture search algorithms such as Group Lasso and Variational Dropout can produce compact neural network architectures that meet performance targets but achieve marginal modeling accuracy when compared to iterative compact architecture search algorithms such as MorphNet and Generative Synthesis. One of the reasons that iterative methods such as MorphNet and Generative Synthesis are able to produce compact deep neural networks with better tradeoff between size, speed, and accuracy is the fact that such methods have much greater freedom to explore design search space of different topologies, thus allowing them to better identify optimal neural network architectures. In particular, Generative Synthesis was able to produce deep neural networks with the best performance trade-offs between size, speed, and accuracy when compared to the other tested methods given that it has the greatest flexibility amongst the methods.  As such, these types of iterative algorithms are more desirable for the compact neural architecture search in practical deep learning development and design workflows.

{\small
	\bibliographystyle{abbrv}
	\bibliography{ccn_style}

\begin{thebibliography}{1}

\bibitem{baker2016designing}
B.~Baker and et~al.
\newblock Designing neural network architectures using reinforcement learning.
\newblock 2016.

\bibitem{deng2009imagenet}
J.~Deng, W.~Dong, R.~Socher, L.-J. Li, K.~Li, and L.~Fei-Fei.
\newblock Imagenet: A large-scale hierarchical image database.
\newblock In {\em 2009 IEEE conference on computer vision and pattern
  recognition}, pages 248--255. Ieee, 2009.

\bibitem{gordon2018morphnet}
A.~Gordon and et~al.
\newblock Morphnet: Fast \& simple resource-constrained structure learning of
  deep networks.
\newblock 2018.

\bibitem{krizhevsky2009learning}
A.~Krizhevsky and G.~Hinton.
\newblock Learning multiple layers of features from tiny images.
\newblock 2009.

\bibitem{molchanov2017variational}
D.~Molchanov and et~al.
\newblock Variational dropout sparsifies deep neural networks.
\newblock 2017.

\bibitem{javad2016evonet}
M.~J. Shafiee and et~al.
\newblock Deep learning with darwin: evolutionary synthesis of deep neural
  networks.
\newblock 2018.

\bibitem{stanley2002evolving}
K.~O. Stanley and R.~Miikkulainen.
\newblock Evolving neural networks through augmenting topologies.
\newblock 2002.

\bibitem{wen2016learning}
W.~Wen and et~al.
\newblock Learning structured sparsity in deep neural networks.
\newblock 2016.

\bibitem{wong2018ferminets}
A.~Wong and et~al.
\newblock Ferminets: Learning generative machines to generate efficient neural
  networks via generative synthesis.
\newblock 2018.

\end{thebibliography}
}
\end{document}